%% file: main.tex
\documentclass[10pt,twocolumn,letterpaper]{article}

\PassOptionsToPackage{table}{xcolor}

\usepackage[pagenumbers]{cvpr}   

\usepackage{subcaption}
\usepackage{booktabs,multirow}

\definecolor{gainGreen}{HTML}{16803C}
\definecolor{gainRed}{HTML}{C0392B}
\newcommand{\gain}[1]{\textcolor{gainGreen}{+#1}}
\newcommand{\loss}[1]{\textcolor{gainRed}{-#1}}
\definecolor{baseGray}{HTML}{A0A0A0}
\newcommand{\base}{\textcolor{baseGray}{--}}


\definecolor{cvprblue}{rgb}{0.21,0.49,0.74}
\usepackage[pagebackref,breaklinks,colorlinks,allcolors=cvprblue]{hyperref}

\def\confName{CVPR}
\def\confYear{2027}

\title{
APS: \underline{A}daptive \underline{P}artitioning-Based Frame
\underline{S}election for Long Video Understanding
}

\author{
Junxiang Wen$^{1}$ \quad
Yinxi Zhao$^{2}$ \quad
Jiawei Chen$^{3}$ \quad
Jingying Hu$^{4}$ \\
Lijun Wang$^{1}$ \quad
Zhen Qin$^{1}$ \quad
Shuiguang Deng$^{1}$ \\[0.5em]
$^{1}$Zhejiang University \qquad
$^{2}$Tsinghua University \\
$^{3}$University of Science and Technology Beijing \\
$^{4}$University of Chinese Academy of Sciences
}

\begin{document}
\maketitle
\input{sec/0_abstract}    
\input{sec/1_intro}
\input{sec/2_related}

\input{sec/3_method}

\input{sec/4_experiments}

\input{sec/5_conclusion}
\section*{LLM Usage Statement}
ChatGPT was used to assist with literature exploration,
language editing, and improving the clarity and presentation of
the manuscript, including figure design and LaTeX formatting.
The authors are responsible for the technical content,
methodological decisions, experimental results, and all claims
presented in this work.
{
    \small
    \bibliographystyle{ieeenat_fullname}
    \bibliography{main}
}

\end{document}

%% file: sec/0_abstract.tex
\begin{abstract}
Recent advances in large vision-language models (LVLMs) have significantly improved video understanding capabilities, yet processing long videos remains challenging due to the substantial computational cost of extensive visual inputs. Frame selection offers an efficient alternative, but many existing region-based methods rely on fixed or predefined temporal partitions, which may result in suboptimal temporal regions, compromising subsequent budget allocation and frame selection. In this work, we propose Adaptive Partitioning-Based Frame Selection (APS), a training-free framework that adapts temporal partitions to local variations in query-frame relevance. Specifically, APS formulates temporal partitioning as a complexity-regularized fitting problem, jointly optimizing partition boundaries and local models through dynamic programming. The resulting adaptive temporal partitions guide frame budget allocation based on query relevance and temporal extent, enabling informative frame selection under limited visual budgets. Extensive experiments on Video-MME, MLVU, and LongVideoBench demonstrate that APS consistently improves performance across diverse LVLM backbones, achieving the highest average gains over uniform sampling among competing methods. Further analyses validate the effectiveness of adaptive partitioning and demonstrate its generalizability across model scales, similarity encoders, and advanced LVLMs.
\end{abstract}

%% file: sec/1_intro.tex
\begin{figure}[t]
    \centering
    \includegraphics[width=0.92\columnwidth]{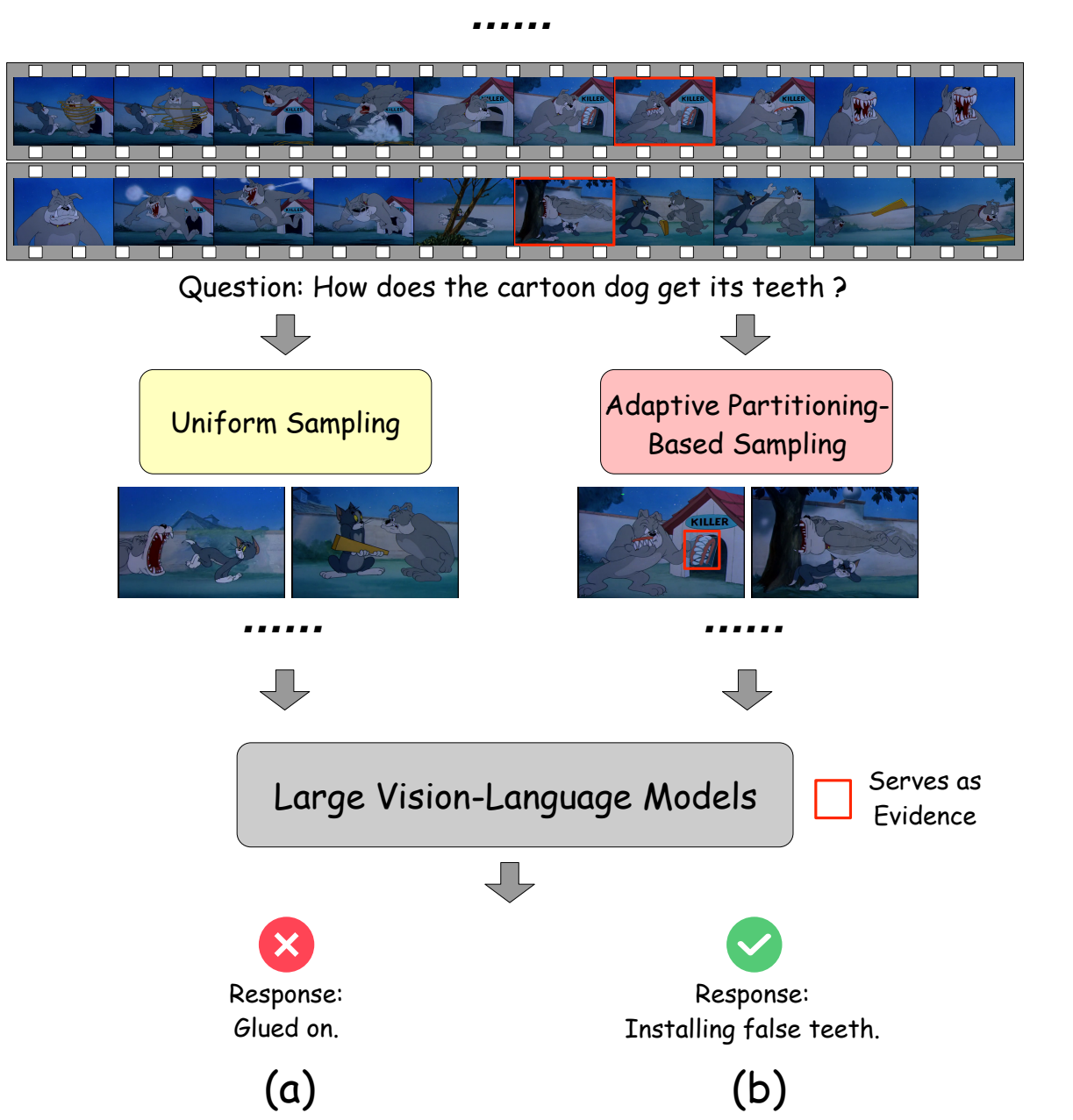}
    \caption{Qualitative comparison of uniform sampling and APS under the same frame budget. (a) Uniform sampling misses critical visual evidence, leading to an incorrect response. (b) APS preserves the relevant evidence through adaptive temporal partitioning, enabling the correct answer. Red boxes highlight the critical visual evidence.}
    \label{fig:intro}
\end{figure}
\section{Introduction}
\label{sec:introduction}

Recent advances in large language models (LLMs) have accelerated the development of large vision-language models (LVLMs), enabling substantial progress in multimodal understanding~\cite{alayrac2022flamingo,liu2023visual,zhang2023video,lin2024video}. Representative LVLMs, including Qwen2.5-VL~\cite{DBLP:journals/corr/abs-2502-13923}, InternVL3~\cite{zhu2025internvl3}, and MiniCPM V4.0~\cite{yao2024minicpm}, have demonstrated strong capabilities in processing visual and video content. Despite these advances, long video understanding remains challenging~\cite{song2024moviechat,chen2025longvila,shu2025video}. Existing LVLMs typically encode video frames into visual tokens, which are subsequently processed by the language model for multimodal reasoning~\cite{li2024llama,zhang2024long,shen2024longvu}. As video duration increases, the growing number of visual tokens incurs increasing computational overhead and may exceed the model's context window, potentially leading to the truncation of visual information~\cite{chen2025longvila,tao2025dycoke,fu2025framefusion,yang2025pvc,liu2025keyframe}. These constraints motivate reducing the visual input consumed by LVLMs while retaining task-relevant evidence.

Frame selection provides a practical solution by retaining a limited set of informative frames for downstream LVLM reasoning. Uniform sampling is a commonly adopted strategy that selects frames at regular temporal intervals to achieve broad video coverage with minimal computational cost. However, its query-agnostic nature may overlook critical evidence, particularly when relevant events are brief or sparsely distributed. Figure~\ref{fig:intro}(a) illustrates this limitation, where uniform sampling misses the critical visual evidence required to answer the query.

To address the limitations of query-agnostic selection, recent methods incorporate query information into frame selection. Some approaches estimate query-frame relevance and combine similarity scores with temporal coverage or multi-resolution selection~\cite{liu2025bolt,zhang2025q}. Others adapt the selection process to query types or employ exploration strategies to identify informative temporal regions~\cite{li2026divide,zhu2026focus}. Temporal organization has also been explored through recursive binary partitioning, predefined temporal segments, and derivative-guided region identification~\cite{tang2025adaptive,peng2026qca,xu2026ridge,Chen_2026_CVPR}. These efforts demonstrate the value of incorporating temporal structure into query-aware frame selection. However, query-frame relevance can exhibit diverse temporal patterns across videos and queries. A particular temporal scheme or region-identification mechanism may merge frames with distinct relevance patterns or fragment coherent high-relevance regions. Determining temporal boundaries that adapt to the underlying relevance structure of each video therefore remains challenging.

Motivated by this challenge and inspired by the adaptive segmentation principle of Bayesian Blocks~\cite{scargle2013studies}, we propose \textbf{APS} (\textbf{Adaptive Partitioning-Based Frame Selection}), a training-free framework for long video frame selection. As illustrated in Figure~\ref{fig:intro}(b), APS preserves the critical visual evidence under the same frame budget through adaptive temporal partitioning. Specifically, APS constructs a temporal relevance sequence from query-frame matching scores obtained using a pretrained vision-language encoder. It then formulates temporal partitioning as a complexity-regularized fitting problem. Dynamic programming jointly determines partition boundaries and local models by balancing fitting error and partition complexity, without relying on a predefined partitioning scheme. The resulting adaptive temporal partitions guide frame budget allocation based on mean query relevance and temporal extent, followed by relevance-based selection within each partition.

Our main contributions are summarized as follows:
\begin{itemize}
    \item We formulate temporal partitioning for long video frame selection as a complexity-regularized fitting problem, enabling adaptive boundary determination without relying on predefined partitioning schemes.

   \item Building on this formulation, we develop APS, a training-free frame selection framework that uses dynamic programming to derive adaptive temporal partitions, followed by inter-partition budget allocation and intra-partition frame selection.

    \item Extensive experiments on three long video benchmarks demonstrate that APS consistently outperforms uniform sampling and achieves the highest average gains across diverse LVLMs. Ablation studies further validate the effectiveness of adaptive temporal partitioning, while additional evaluations demonstrate its generalization across model scales, similarity encoders, and advanced LVLMs. Our code is publicly available on \href{https://github.com/Baibaione/APS}{GitHub}.
\end{itemize}

%% file: sec/2_related.tex
\section{Related Work}
\label{sec:related_work}

\paragraph{Long-Context Modeling and Visual Compression.}

Recent advances in large vision-language models (LVLMs) have improved video understanding, yet processing long videos remains computationally demanding due to extensive visual inputs. Existing approaches address this challenge primarily through long-context modeling and visual representation compression. LongVILA~\cite{chen2025longvila}, LongVA~\cite{zhang2024long}, and Video-XL~\cite{shu2025video} extend the context capacity of LVLMs to accommodate longer video sequences, while MovieChat~\cite{song2024moviechat} employs memory mechanisms to retain important temporal information. Apollo~\cite{zohar2025apollo} further investigates video-specific modeling choices. Alternatively, visual compression methods reduce the number of tokens required to represent video content. LongVU~\cite{shen2024longvu} introduces adaptive spatiotemporal compression, while DyCoke~\cite{tao2025dycoke}, FrameFusion~\cite{fu2025framefusion}, and PVC~\cite{yang2025pvc} exploit visual redundancy to improve processing efficiency. LLaMA-VID~\cite{li2024llama}, Dynamic-VLM~\cite{wang2025dynamic}, and One Token per Highly Selective Frame~\cite{zhang2026one} further investigate compact or adaptive visual representations. Unlike approaches that extend context capacity or compress visual tokens, frame selection directly determines which frames are retained for downstream reasoning.

\begin{figure*}[t]
    \centering
    \includegraphics[width=\textwidth]{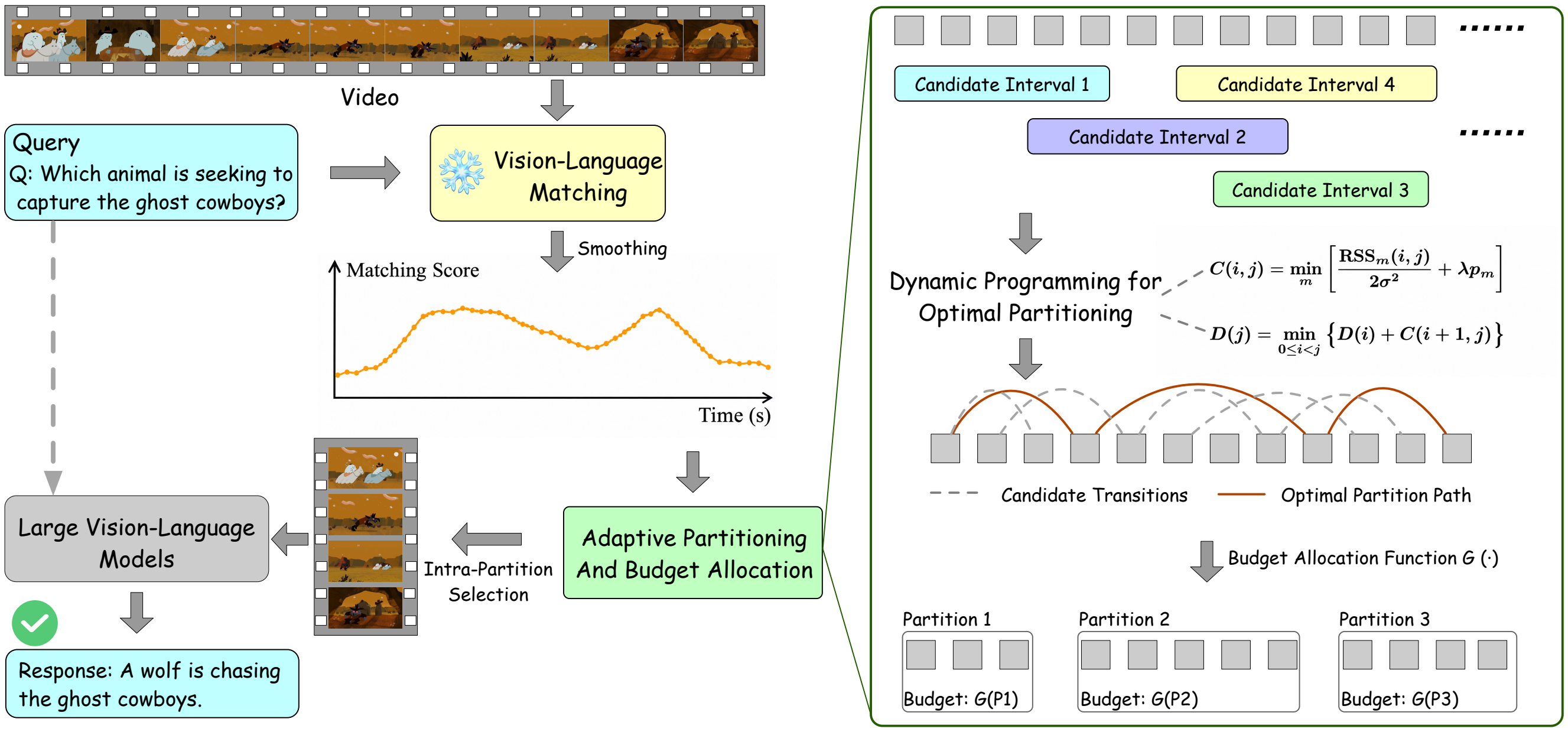}
    \caption{Overview of APS. Given a video and a textual query, APS constructs a temporal relevance sequence, determines adaptive temporal partitions through complexity-regularized dynamic programming, and allocates the frame budget across partitions for keyframe selection.}
    \label{fig:framework}
\end{figure*}

\paragraph{Frame Selection for Long Video Understanding.}

Frame selection aims to identify informative frames under constrained visual budgets by balancing query relevance, temporal coverage, and visual diversity. Similarity-based approaches use query--frame matching to guide frame selection, with BOLT~\cite{liu2025bolt} exploring training-free sampling strategies and Q-Frame~\cite{zhang2025q} combining query-aware selection with multi-resolution adaptation. AKS~\cite{tang2025adaptive} jointly considers query relevance and temporal coverage through recursive binary partitioning. MDP3~\cite{sun2025mdp} incorporates query-conditioned similarity, list-wise diversity, and sequential dependencies through determinantal point processes and a Markov decision process, while MaxInfo~\cite{li2026maxinfo} selects representative frames based on a maximum-volume criterion. Other methods adapt frame selection to query characteristics or informative temporal regions: DIG~\cite{li2026divide} adjusts selection according to query types, whereas FOCUS~\cite{zhu2026focus} employs exploration to identify informative regions. Beyond training-free approaches, reinforcement learning has been explored to learn frame selection policies~\cite{qin2026efficient}, while VideoBrain~\cite{zou2026videobrain} learns adaptive visual information acquisition strategies. VideoZoomer~\cite{ding2026videozoomer} and FrameThinker~\cite{he2026framethinker} further investigate iterative visual evidence acquisition during video reasoning.

Existing frame selection methods employ different strategies to organize query-relevant temporal information. AKS~\cite{tang2025adaptive} uses recursive binary partitioning to balance query relevance and temporal coverage, while QCA~\cite{peng2026qca} allocates frame budgets across predefined uniform segments. RIDGE~\cite{xu2026ridge} identifies informative temporal regions through derivative-guided analysis of query--frame relevance sequences. These methods exploit temporal structures from different perspectives, but their partitioning or region-identification strategies may not consistently adapt to diverse temporal patterns of query relevance across videos. In contrast, APS formulates temporal partitioning as a complexity-regularized fitting problem, jointly determining partition boundaries and local model types through dynamic programming without requiring predefined partitioning schemes. This enables APS to adapt temporal partitions to local relevance structures and guide subsequent frame budget allocation.

%% file: sec/3_method.tex
\section{Method}
\label{sec:method}

\subsection{Overview}
\label{sec:overview}

Given a long video and a textual query, APS selects a compact set of query-relevant frames under a fixed visual budget. As illustrated in Figure~\ref{fig:framework}, the framework consists of three stages. First, a frozen vision-language encoder generates a temporal sequence of query--frame relevance scores. Second, complexity-regularized local fitting determines adaptive temporal partitions and their associated models via dynamic programming. Finally, frame budgets are allocated across partitions, followed by relevance-based frame selection within each partition.

\subsection{Query-Frame Relevance Estimation}
\label{sec:relevance}

Let $\mathcal{V}=\{f_i\}_{i=1}^{N}$ be the candidate frames sampled uniformly from a video, $q$ the input query, and $K$ the frame budget of the downstream video-language model. A frozen vision-language encoder $g$~\cite{li2022blip} assigns each frame a query-relevance score:
\begin{equation}
s_i=g(f_i,q), \qquad i=1,\ldots,N.
\label{eq:relevance_score}
\end{equation}

To obtain comparable scores within each video, we apply min-max normalization:
\begin{equation}
x_i=
\begin{cases}
\dfrac{s_i-s_{\min}}{s_{\max}-s_{\min}},
& s_{\max}-s_{\min}\geq\varepsilon,\\[6pt]
0 & \text{otherwise}
\end{cases}
\label{eq:score_normalization}
\end{equation}
where $s_{\min}$ and $s_{\max}$ are the minimum and maximum scores, and $\varepsilon>0$ prevents numerical instability when the score range is negligible. We then suppress short-range fluctuations using a moving average:
\begin{equation}
y_i=\frac{1}{|\mathcal{N}_i|}
\sum_{t\in\mathcal{N}_i}x_t,
\label{eq:score_smoothing}
\end{equation}
where $\mathcal{N}_i=\{t:1\leq t\leq N,\ |t-i|\leq h\}$ and $h$ is the smoothing half-width. The smoothed sequence $\mathbf{y}$ determines temporal partitions, whereas the normalized scores $\mathbf{x}$ are used for budget allocation and final frame selection.

\subsection{Adaptive Temporal Partitioning}
\label{sec:temporal_partitioning}

To realize adaptive temporal partitioning for query-aware frame selection, we formulate the problem as complexity-regularized piecewise regression, jointly optimizing interval boundaries and local model types to approximate the query-frame relevance trajectory.

\paragraph{Local Parametric Modeling.}
For a candidate interval $I=[l,r]$ of length $L=r-l+1$, we consider constant and linear models:
\begin{equation}
\begin{aligned}
\mathcal{M}_0 &: \quad \hat{y}_t=a,\\
\mathcal{M}_1 &: \quad \hat{y}_t=a+bt,
\end{aligned}
\label{eq:local_parametric_models}
\end{equation}
which represent approximately stable relevance levels and local trends, respectively. The fitting error of each model $m$ is measured by its least-squares residual:
\begin{equation}
\operatorname{RSS}_m(I)
=
\min_{\boldsymbol{\theta}_m}
\sum_{t=l}^{r}
\left(y_t-\hat{y}_t^{(m)}\right)^2,
\label{eq:local_rss}
\end{equation}
where $\boldsymbol{\theta}_m$ denotes the model parameters.

\paragraph{Complexity-Regularized Modeling.}
Minimizing residual error alone can favor overly flexible fits and fragmented partitions. We therefore assign each interval the penalized cost
\begin{equation}
C(I)=
\min_{m\in\mathcal{M}(I)}
\left[
\frac{\operatorname{RSS}_m(I)}{2\sigma^2}
+\lambda p_m
\right],
\label{eq:interval_cost}
\end{equation}
where $\mathcal{M}(I)$ is the set of admissible local models, $p_m$ their parameter count ($p_0=1$ and $p_1=2$), $\sigma$ the relevance noise scale, and $\lambda$ the complexity coefficient. We estimate $\sigma$ robustly from the first differences $d_i=x_{i+1}-x_i$:
\begin{equation}
\sigma=
\max\left(
\frac{1.4826}{\sqrt{2}}
\operatorname{MAD}(\{d_i\}),
\sigma_{\min}
\right),
\label{eq:noise_scale}
\end{equation}
where $\operatorname{MAD}$ is the median absolute deviation and $\sigma_{\min}>0$ prevents a vanishing scale. Using a BIC-inspired penalty~\cite{schwarz1978estimating}, we set
\begin{equation}
\lambda=\frac{\eta}{2}\log N,
\label{eq:complexity_coefficient}
\end{equation}
with $\eta$ controlling regularization strength. The penalty balances fitting accuracy against model complexity, discouraging unnecessarily complex local models and excessive partitioning.

\paragraph{Dynamic Programming for Temporal Partitioning.}
Let $\mathcal{P}=\{I_1,\ldots,I_J\}$ be a partition into contiguous, non-overlapping intervals. We determine both the boundaries and the number of partitions by minimizing the total interval cost:
\begin{equation}
\mathcal{P}^{*}
=
\arg\min_{\mathcal{P}}
\sum_{j=1}^{J}C(I_j).
\label{eq:optimal_partition}
\end{equation}

The additive objective admits dynamic programming. Let $D(t)$ be the minimum cost of partitioning the first $t$ frames, with $D(0)=0$. Then
\begin{equation}
D(t)=
\min_{k\in\mathcal{A}(t)}
\left\{
D(k)+C([k+1,t])
\right\},
\label{eq:dp_partition}
\end{equation}
where $\mathcal{A}(t)$ contains feasible preceding boundaries. Backtracking from $D(N)$ recovers the optimal partition and its associated local model for each interval, without fixing $J$ in advance.

\subsection{Partition-Guided Frame Selection}
\label{sec:frame_selection}

\paragraph{Inter-Partition Budget Allocation.}
For each partition $I_j$, we compute its mean relevance $\mu_j$ and temporal length $L_j$:
\begin{equation}
\mu_j=\frac{1}{L_j}\sum_{i\in I_j}x_i,
\qquad L_j=|I_j|.
\label{eq:partition_statistics}
\end{equation}
Each partition is assigned a weight:
\begin{equation}
w_j=(\mu_j+\epsilon_w)^\alpha L_j^\beta,
\label{eq:partition_weight}
\end{equation}
where $\alpha$ and $\beta$ control the contributions of query relevance and temporal extent, respectively, and $\epsilon_w>0$ ensures positive weights.

We define a capacity-constrained allocation function $G$ that maps partition weights $\mathbf{w}$, lengths $\mathbf{L}$, and the total budget $K$ to an integer budget vector $\mathbf{b}$:
\begin{equation}
\mathbf{b}=G(\mathbf{w},\mathbf{L},K),
\label{eq:budget_function}
\end{equation}
where $\mathbf{w}=(w_1,\ldots,w_J)$, $\mathbf{L}=(L_1,\ldots,L_J)$, and $\mathbf{b}=(b_1,\ldots,b_J)$.

When $K<J$, one frame is assigned to each of the $K$ highest-weight partitions:
\begin{equation}
b_j=\mathbb{I}\left[\operatorname{rank}(w_j)\leq K\right],
\label{eq:budget_priority}
\end{equation}
where $\mathbb{I}[\cdot]$ is the indicator function, and ties are resolved by temporal order.

When $K\geq J$, each partition initially receives one frame, and the remaining budget is distributed proportionally to the partition weights:
\begin{equation}
\widetilde{b}_j
=1+(K-J)\frac{w_j}{\sum_{k=1}^{J}w_k}.
\label{eq:budget_proportion}
\end{equation}
The target allocations $\widetilde{b}_j$ are converted to integer budgets through capacity-constrained redistribution. Assuming $K\leq N$, the final allocation satisfies
\begin{equation}
\sum_{j=1}^{J}b_j=K,
\qquad 0\leq b_j\leq L_j.
\label{eq:allocation_constraint}
\end{equation}

\paragraph{Intra-Partition Frame Selection.}
Within each partition, we select the $b_j$ frames with the highest normalized relevance scores:
\begin{equation}
\mathcal{S}_j^{*}
=
\arg\max_{\substack{
\mathcal{S}_j\subseteq I_j\\
|\mathcal{S}_j|=b_j
}}
\sum_{i\in\mathcal{S}_j}x_i.
\label{eq:intra_partition_topk}
\end{equation}
The final selection is the union of the partition-wise subsets:
\begin{equation}
\mathcal{S}^{*}
=
\bigcup_{j=1}^{J}\mathcal{S}_j^{*}.
\label{eq:selected_frames}
\end{equation}
Selected frames are arranged in chronological order before being passed to the downstream video-language model.

%% file: sec/4_experiments.tex
\begin{table*}[t]
\centering
\caption{Comparison with training-free frame selection methods on Video-MME, MLVU, and LongVideoBench. $\Delta$ is the accuracy gain (percentage points) over uniform sampling, averaged across datasets in the final column. Best and second-best accuracies are shown in \textbf{bold} and \underline{underlined}, respectively. The highest gains are also highlighted in \textbf{bold}.}
\label{tab:main_results}
\setlength{\tabcolsep}{5pt}
\renewcommand{\arraystretch}{1.12}
\begin{tabular}{llcccccccc}
\toprule
\multirow{2}{*}{Model} & \multirow{2}{*}{Method} & \multirow{2}{*}{Frames}
& \multicolumn{2}{c}{Video-MME}
& \multicolumn{2}{c}{MLVU}
& \multicolumn{2}{c}{LongVideoBench}
& \multirow{2}{*}{Avg. $\Delta$} \\
\cmidrule(lr){4-5}\cmidrule(lr){6-7}\cmidrule(lr){8-9}
& & & Acc. & $\Delta$ & Acc. & $\Delta$ & Acc. & $\Delta$ & \\
\midrule
\multirow{7}{*}{Qwen2.5-VL-7B}
& Uniform & 32 & 59.7 & \base & 60.7 & \base & 58.0 & \base & \base \\
& AKS {\scriptsize [CVPR'25]} & 32 & 62.0 & \gain{2.3} & 65.2 & \gain{4.5} & \underline{63.2} & \gain{5.2} & \gain{4.00} \\
& DIG {\scriptsize [CVPR'26]} & 32 & 60.4 & \gain{0.7} & \underline{68.2} & \gain{7.5} & 61.5 & \gain{3.5} & \gain{3.90} \\
& QCA {\scriptsize [ECCV'26]} & 32 & \underline{62.1} & \gain{2.4} & 66.9 & \gain{6.2} & 61.5 & \gain{3.5} & \gain{4.03} \\
& FOCUS {\scriptsize [ICLR'26]} & 32 & 60.4 & \gain{0.7} & 68.0 & \gain{7.3} & 61.2 & \gain{3.2} & \gain{3.73} \\
& RIDGE {\scriptsize [EMNLP'26]} & 32 & \underline{62.1} & \gain{2.4} & 67.3 & \gain{6.6} & 62.7 & \gain{4.7} & \underline{\gain{4.57}} \\
\rowcolor{orange!25}
& \textbf{APS} & 32 & \textbf{62.3} & \textbf{\gain{2.6}} & \textbf{69.6} & \textbf{\gain{8.9}} & \textbf{63.8} & \textbf{\gain{5.8}} & \textbf{\gain{5.77}} \\
\midrule
\multirow{7}{*}{InternVL3-8B}
& Uniform & 32 & 64.6 & \base & 70.2 & \base & 59.3 & \base & \base \\
& AKS {\scriptsize [CVPR'25]} & 32 & 66.2 & \gain{1.6} & 74.1 & \gain{3.9} & 61.7 & \gain{2.4} & \gain{2.63} \\
& DIG {\scriptsize [CVPR'26]} & 32 & 65.0 & \gain{0.4} & 73.2 & \gain{3.0} & 62.4 & \gain{3.1} & \gain{2.17} \\
& QCA {\scriptsize [ECCV'26]} & 32 & \underline{66.4} & \gain{1.8} & 74.5 & \gain{4.3} & 62.3 & \gain{3.0} & \underline{\gain{3.03}} \\
& FOCUS {\scriptsize [ICLR'26]} & 32 & 63.9 & \loss{0.7} & 72.7 & \gain{2.5} & 61.6 & \gain{2.3} & \gain{1.37} \\
& RIDGE {\scriptsize [EMNLP'26]} & 32 & 65.5 & \gain{0.9} & \underline{75.0} & \gain{4.8} & \underline{62.5} & \gain{3.2} & \gain{2.97} \\
\rowcolor{orange!25}
& \textbf{APS} & 32 & \textbf{66.9} & \textbf{\gain{2.3}} & \textbf{75.2} & \textbf{\gain{5.0}} & \textbf{63.6} & \textbf{\gain{4.3}} & \textbf{\gain{3.87}} \\
\midrule
\multirow{7}{*}{MiniCPM V4.0-4B}
& Uniform & 16 & 56.4 & \base & 53.0 & \base & 54.4 & \base & \base \\
& AKS {\scriptsize [CVPR'25]} & 16 & \textbf{58.4} & \textbf{\gain{2.0}} & 60.5 & \gain{7.5} & 57.4 & \gain{3.0} & \underline{\gain{4.17}} \\
& DIG {\scriptsize [CVPR'26]} & 16 & 57.5 & \gain{1.1} & 59.3 & \gain{6.3} & 57.1 & \gain{2.7} & \gain{3.37} \\
& QCA {\scriptsize [ECCV'26]} & 16 & \underline{57.6} & \gain{1.2} & 60.8 & \gain{7.8} & 56.1 & \gain{1.7} & \gain{3.57} \\
& FOCUS {\scriptsize [ICLR'26]} & 16 & 54.9 & \loss{1.5} & \underline{61.5} & \gain{8.5} & \textbf{58.1} & \textbf{\gain{3.7}} & \gain{3.57} \\
& RIDGE {\scriptsize [EMNLP'26]} & 16 & 55.2 & \loss{1.2} & 56.1 & \gain{3.1} & \underline{57.9} & \gain{3.5} & \gain{1.80} \\
\rowcolor{orange!25}
& \textbf{APS} & 16 & \underline{57.6} & \gain{1.2} & \textbf{61.8} & \textbf{\gain{8.8}} & \textbf{58.1} & \textbf{\gain{3.7}} & \textbf{\gain{4.57}} \\
\bottomrule
\end{tabular}
\end{table*}

\begin{table}[t]
\centering
\small
\caption{Ablation results on Video-MME and LongVideoBench using Qwen2.5-VL-7B with 32 frames. Equal Partition uses uniform boundaries while preserving the partition count. LVB denotes LongVideoBench.}
\label{tab:ablation}
\setlength{\tabcolsep}{7pt}
\begin{tabular}{lcc}
\toprule
\textbf{Variant} & \textbf{Video-MME} & \textbf{LVB} \\
\midrule
Equal Partition        & 61.59 & 62.23 \\
Constant-Only          & 61.93 & 63.05 \\
Linear-Only            & 62.07 & 62.83 \\
w/o Complexity Penalty & 61.81 & 62.83 \\
Relevance-only Allocation & 61.93 & 63.50 \\
\midrule
\textbf{APS (Full)}    & \textbf{62.33} & \textbf{63.80} \\
\bottomrule
\end{tabular}
\end{table}
\section{Experiments}
\label{sec:experiments}

\subsection{Experimental Setup}
\label{sec:experimental_setup}

\paragraph{Benchmarks and Backbones.}
We evaluate APS on Video-MME~\cite{fu2025video}, MLVU~\cite{zhou2025mlvu}, and LongVideoBench~\cite{wu2024longvideobench}, which collectively cover diverse video durations, temporal reasoning, and long-context question answering. Subtitles are not used. To examine whether the frame selection strategy transfers across different LVLMs, we use Qwen2.5-VL-7B~\cite{DBLP:journals/corr/abs-2502-13923}, InternVL3-8B~\cite{zhu2025internvl3}, and MiniCPM V4.0-4B~\cite{yao2024minicpm} as downstream backbones.

\paragraph{Comparison Protocol.}
We compare APS against uniform sampling and five training-free frame selection methods: AKS~\cite{tang2025adaptive}, DIG~\cite{li2026divide}, QCA~\cite{peng2026qca}, FOCUS~\cite{zhu2026focus}, and RIDGE~\cite{xu2026ridge}. All methods use the same frame budget for a given downstream backbone: 32 frames for Qwen2.5-VL-7B and InternVL3-8B, and 16 for MiniCPM V4.0-4B. APS constructs its candidate pool by sampling the original video at 1 FPS and computes query--frame relevance with a frozen BLIP~\cite{li2022blip} encoder; competing methods retain their respective feature-encoder configurations. We report multiple-choice question-answering accuracy (\%) and its absolute improvement over uniform sampling, $\Delta$, in percentage points. Avg.~$\Delta$ is the average improvement across the three benchmarks for each backbone.

\paragraph{Implementation Details.}
APS uses a complexity regularization coefficient of $\eta=0.5$ and a relevance weight of $\alpha=1.0$. The temporal-length weight is $\beta=0.4$ on Video-MME and MLVU, and $\beta=0.6$ on LongVideoBench. Unless stated otherwise, experiments retain these settings. Additional analyses vary the Qwen2.5-VL model size, similarity encoder, and frame budget while keeping the remaining protocol fixed. 
For the additional experiments with GLM-5.3-Flash~\cite{zeng2026glm}
and DeepSeek-V4.1-Flash~\cite{xu2026deepseek},
both models are evaluated in low thinking mode. All experiments are conducted on a platform with 8 GPUs (64 GB memory per GPU and 280 TFLOPS peak BF16 throughput per GPU). We utilize the LMMs-Eval toolkit~\cite{zhang2025lmms} for downstream LVLM evaluation.

\subsection{Main Results}
\label{sec:main_results}

\paragraph{Comparison with Existing Methods.}
Table~\ref{tab:main_results} presents a comprehensive comparison of APS with existing training-free frame selection methods across three LVLM backbones and three long-video benchmarks. APS demonstrates strong adaptability across different benchmarks and model architectures, consistently outperforming uniform sampling in all nine settings, achieving the best or tied-best accuracy in eight, and obtaining the highest average gain for each backbone. This consistent performance across diverse evaluation settings highlights the effectiveness of APS without relying on a specific model architecture or benchmark. Moreover, APS maintains its advantage even with the smaller MiniCPM V4.0-4B backbone under a reduced frame budget of 16, demonstrating its effectiveness when visual inputs are constrained. Overall, APS provides an effective training-free solution for long-video understanding through adaptive temporal partitioning guided by query relevance.

\subsection{Ablation and Parameter Analysis}
\label{sec:ablation_analysis}

\paragraph{Ablation of Key Components.}
We conduct ablation studies to evaluate the key design choices in APS. As shown in Table~\ref{tab:ablation}, APS consistently outperforms all variants across both benchmarks. Equal Partition replaces adaptive boundaries with uniformly spaced ones while preserving the partition count, highlighting the importance of aligning temporal boundaries with local relevance patterns. Constant-Only and Linear-Only restrict each interval to a single model type. The former may overlook gradual relevance trends, whereas the latter may capture minor fluctuations in stable regions, demonstrating the benefit of combining both models. Removing complexity regularization eliminates the penalty on model complexity, making the partitioning process more susceptible to local fluctuations. Finally, Relevance-only Allocation distributes frame budgets according to partition relevance without considering temporal length. Its lower performance suggests that temporal extent provides complementary information to query relevance for budget allocation.

\begin{figure*}[htbp]
    \centering
    \captionsetup[subfigure]{justification=centering,singlelinecheck=false}

    \begin{subfigure}[t]{0.24\textwidth}
        \centering
        \includegraphics[width=\linewidth]{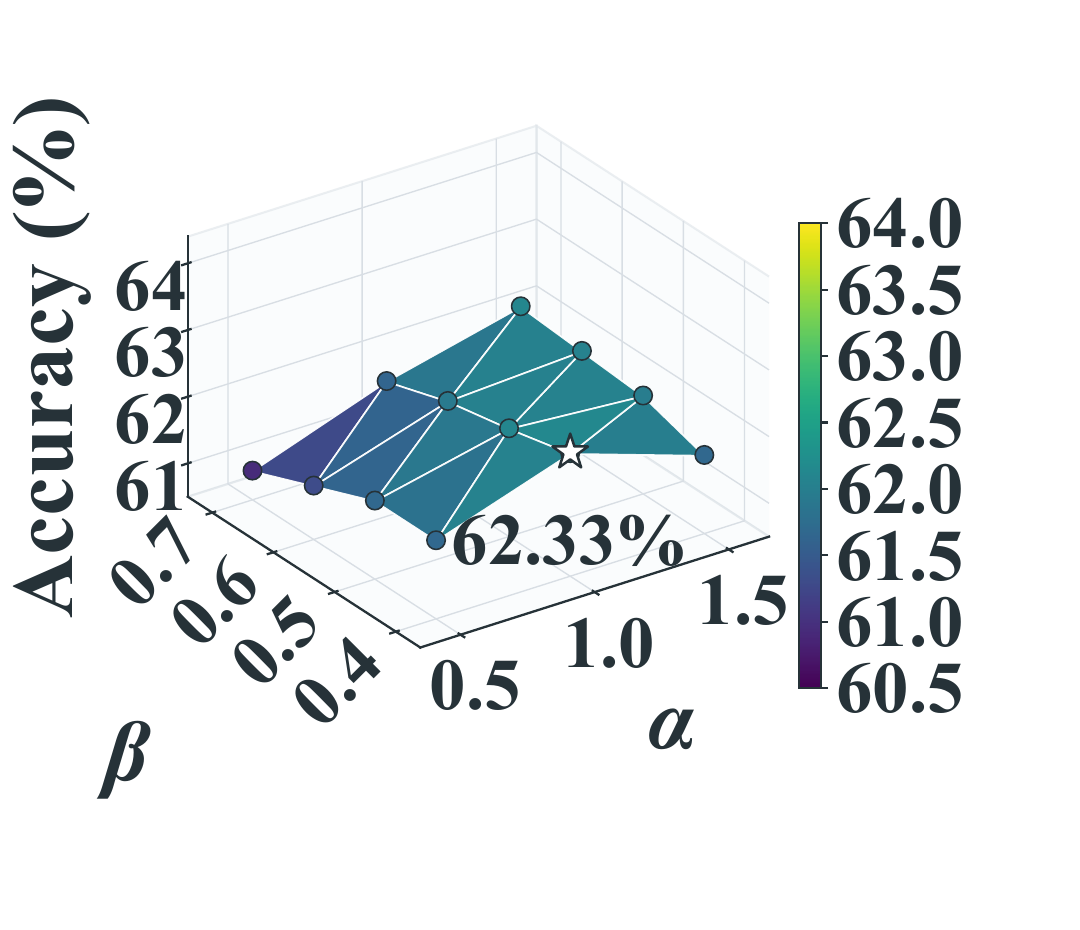}
        \caption{Sensitivity on Video-MME.}
        \label{fig:sensitivity_videomme}
    \end{subfigure}\hfill%
    \begin{subfigure}[t]{0.24\textwidth}
        \centering
        \includegraphics[width=\linewidth]{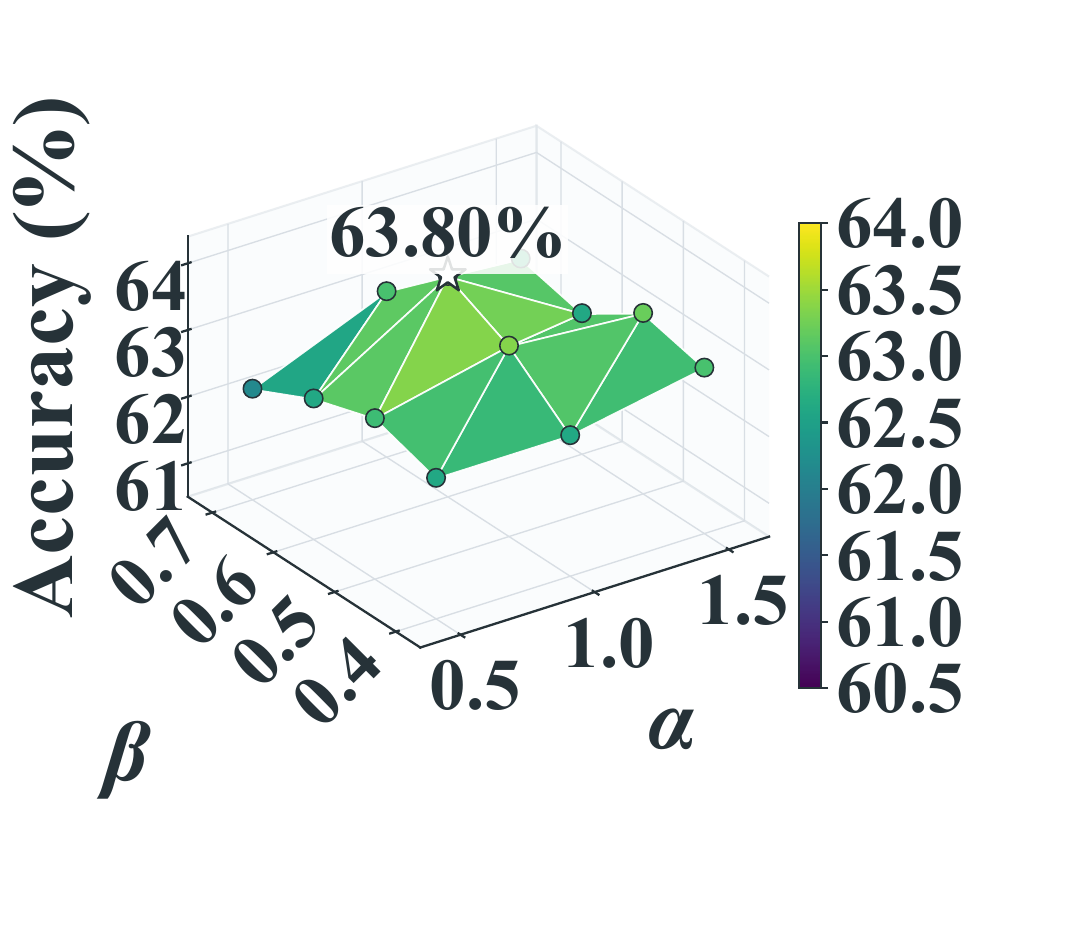}
        \caption{Sensitivity on LongVideoBench.}
        \label{fig:sensitivity_lvb}
    \end{subfigure}\hfill%
    \begin{subfigure}[t]{0.24\textwidth}
        \centering
        \includegraphics[width=\linewidth]{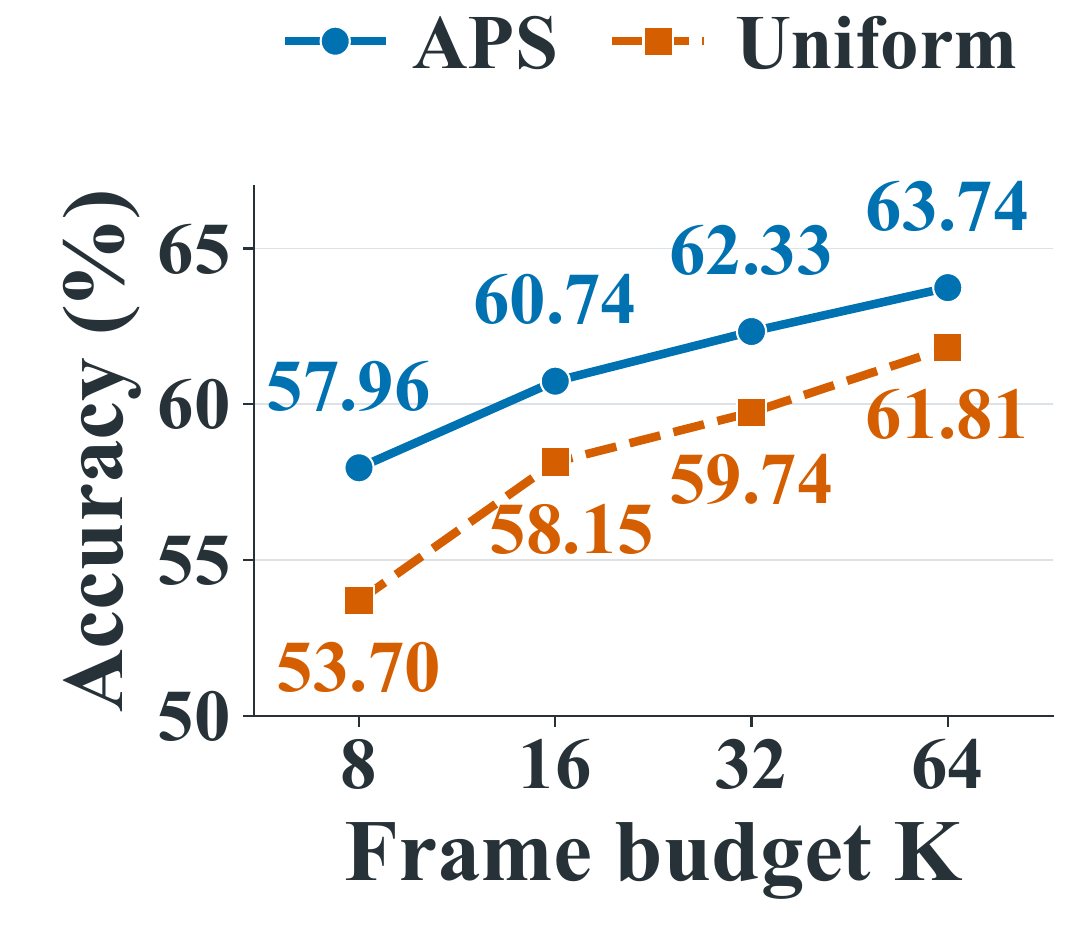}
        \caption{Frame budget on Video-MME.}
        \label{fig:budget_videomme}
    \end{subfigure}\hfill%
    \begin{subfigure}[t]{0.24\textwidth}
        \centering
        \includegraphics[width=\linewidth]{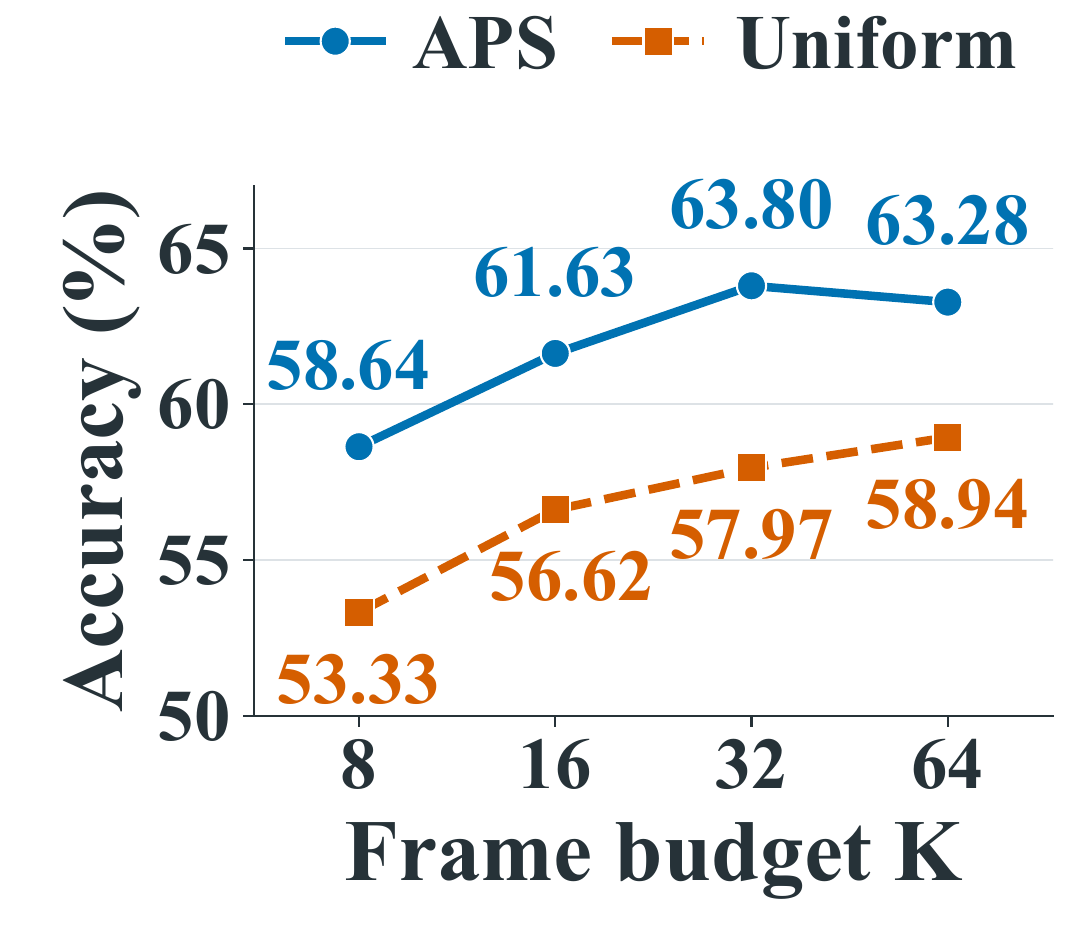}
        \caption{Frame budget on LongVideoBench.}
        \label{fig:budget_lvb}
    \end{subfigure}

    \caption{Parameter sensitivity and frame budget analysis of APS.
    (a)--(b) examine the sensitivity to budget allocation parameters
    $\alpha$ and $\beta$.
    (c)--(d) compare APS with uniform sampling under different frame budgets.
    All experiments use Qwen2.5-VL-7B with BLIP similarity scores.}
    \label{fig:parameter_budget}
\end{figure*}

\paragraph{Sensitivity of Budget Allocation.}
APS allocates frame budgets according to partition-level mean query relevance and temporal extent, weighted by $\alpha$ and $\beta$, respectively. Figure~\ref{fig:parameter_budget}(a)--(b) shows that both benchmarks achieve their highest accuracy at $\alpha=1.0$, while the optimal $\beta$ varies across datasets. Performance remains competitive under neighboring configurations, suggesting that budget allocation is robust to moderate parameter changes.

\begin{table}[t]
\centering
\small
\caption{Sensitivity to the complexity regularization coefficient $\eta$ using InternVL3-8B with 32 frames.}
\label{tab:penalty_sensitivity}
\setlength{\tabcolsep}{5pt}
\begin{tabular}{lccc}
\toprule
\textbf{$\eta$} & \textbf{Video-MME} & \textbf{MLVU} & \textbf{LVB} \\
\midrule
0.25  & 66.5 & \textbf{75.9} & 63.5 \\
0.50  & 66.9 & 75.2 & 63.6 \\
0.75  & \textbf{67.0} & 75.0 & \textbf{63.9} \\
1.00  & 66.4 & 74.9 & 63.6 \\
1.25  & 66.6 & 74.7 & 63.8 \\
\bottomrule
\end{tabular}
\end{table}
\paragraph{Sensitivity to Complexity Regularization.}
Table~\ref{tab:penalty_sensitivity} examines the effect of the complexity regularization coefficient $\eta$ using InternVL3-8B with 32 frames. APS maintains stable performance across the tested range, indicating robustness to regularization strength. Notably, the standard BIC penalty corresponds to $\eta=1$, while the results in our frame selection setting generally favor smaller values, suggesting that a moderately relaxed complexity penalty may better balance local fitting accuracy and partition complexity for modeling query-frame relevance.

\begin{figure*}[t]
    \centering
    \includegraphics[width=0.9\textwidth]{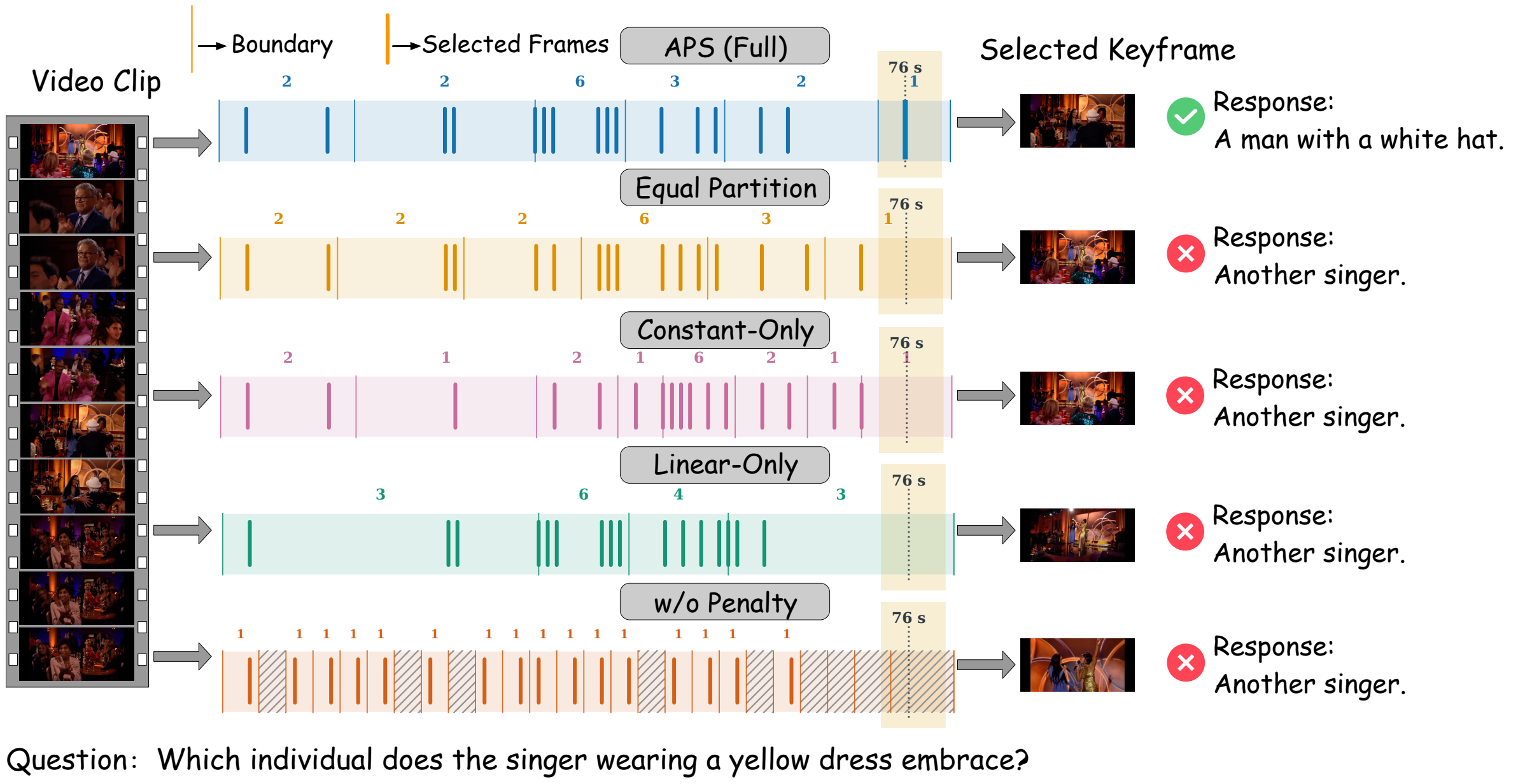}
    \caption{Qualitative comparison of APS and four ablation variants on Video-MME under a fixed 16-frame budget. Timelines show temporal partitions, allocated budgets, and selected frames. The highlighted region contains the critical evidence at 76s, while hatched regions indicate zero-budget segments. Only APS preserves the critical evidence and answers correctly.}
    \label{fig:qualitative}
\end{figure*}

\paragraph{Qualitative Visualization.}
Figure~\ref{fig:qualitative} compares APS with four ablation variants on a Video-MME example under the same frame budget. APS preserves the critical frame at 76s and answers correctly, whereas all variants miss it. Equal Partition uses uniform boundaries that may group critical evidence with less informative content, weakening its representation during selection. Constant-Only cannot effectively capture evolving relevance trends, favoring more fragmented partitions, while Linear-Only incurs higher model complexity penalties that can favor fewer partitions and merge distinct relevance patterns. Without complexity regularization, excessive fragmentation produces 25 segments, leaving the evidence-containing segment with zero frame budget. This illustrates that finer temporal partitioning does not necessarily improve evidence coverage under a fixed frame budget. These differences illustrate how partition structure influences frame allocation and evidence preservation under the same visual budget.

\subsection{Generalization and Budget Efficiency}
\label{sec:further_analysis}
\begin{table}[t]
\centering
\caption{Effect of query--frame similarity encoders with Qwen2.5-VL-7B and 32 frames. Uniform sampling is shared across encoders; $\Delta$ is the gain over uniform.}
\label{tab:encoder_ablation}
\setlength{\tabcolsep}{6pt}
\renewcommand{\arraystretch}{1.12}
\begin{tabular}{lcccc}
\toprule
\multirow{2}{*}{Encoder}
& \multicolumn{2}{c}{Video-MME}
& \multicolumn{2}{c}{LongVideoBench} \\
\cmidrule(lr){2-3}\cmidrule(lr){4-5}
& Acc. & $\Delta$ & Acc. & $\Delta$ \\
\midrule
Uniform & 59.74 & \base & 57.97 & \base \\
\midrule
BLIP   & \textbf{62.33} & \gain{2.59} & \textbf{63.80} & \gain{5.83} \\
CLIP   & 62.26 & \gain{2.52} & 62.23 & \gain{4.26} \\
BLIP-2 & 61.19 & \gain{1.45} & 63.13 & \gain{5.16} \\
\bottomrule
\end{tabular}
\end{table}

\begin{table}[t]
\centering
\caption{Effect of backbone scale with Qwen2.5-VL, 32 frames, and BLIP scores. Avg. is the mean accuracy across Video-MME and MLVU.}
\label{tab:model_scale}
\setlength{\tabcolsep}{5pt}
\renewcommand{\arraystretch}{1.1}
\small
\begin{tabular}{llccc}
\toprule
Scale & Method & Video-MME & MLVU & Avg. \\
\midrule
\multirow{2}{*}{3B}
& Uniform & 57.74 & 59.43 & 58.59 \\
& APS & \textbf{59.30} & \textbf{68.77} & \textbf{64.04} \\
\midrule
\multirow{2}{*}{7B}
& Uniform & 59.74 & 60.67 & 60.21 \\
& APS & \textbf{62.33} & \textbf{69.64} & \textbf{65.99} \\
\midrule
\multirow{2}{*}{32B}
& Uniform & 62.78 & 61.09 & 61.94 \\
& APS & \textbf{63.30} & \textbf{72.31} & \textbf{67.81} \\
\bottomrule
\end{tabular}
\end{table}

\paragraph{Frame Budget.}
We analyze the impact of frame budgets ranging from 8 to 64 using Qwen2.5-VL-7B with BLIP scores. Figure~\ref{fig:parameter_budget}(c)--(d) shows that APS consistently outperforms uniform sampling across all budgets. Notably, APS with 16 frames surpasses uniform sampling with 32 frames on both benchmarks, highlighting the importance of informative frame selection under constrained budgets. On LongVideoBench, APS peaks at 32 frames and declines slightly at 64, whereas uniform sampling continues to improve. This suggests that additional frames may offer diminishing benefits or introduce redundant visual information once relevant evidence has been captured~\cite{wang2026less}.

\paragraph{Similarity Encoder.}
We examine APS with three pretrained vision-language encoders: BLIP~\cite{li2022blip}, CLIP~\cite{radford2021learning}, and BLIP-2~\cite{li2023blip}, using Qwen2.5-VL-7B with 32 frames. Table~\ref{tab:encoder_ablation} shows that APS outperforms uniform sampling with all three encoders on both benchmarks. BLIP achieves the highest accuracy, while CLIP performs better than BLIP-2 on Video-MME but worse on LongVideoBench. These results indicate that APS is compatible with different similarity representations, although encoder choice influences selection performance.

\paragraph{Backbone Scale.}
To examine the effect of model capacity, we compare APS across three Qwen2.5-VL scales (3B, 7B, and 32B) under a fixed 32-frame budget. Table~\ref{tab:model_scale} shows consistent improvements over uniform sampling at all scales on both benchmarks. These results indicate that APS remains effective across backbones with different capacities. Notably, APS achieves an average improvement of 5.87 percentage points with the 32B backbone, demonstrating its continued benefit for larger models.


\paragraph{Computational Efficiency.}
Table~\ref{tab:efficiency} compares the runtime of APS, AKS, and QCA across processing stages. Feature extraction dominates the total runtime, while adaptive partitioning accounts for approximately 0.3\% of APS processing time. Since APS and AKS share the same feature extraction cost, their total runtimes differ by only 0.045 seconds per sample. Budget allocation and frame selection introduce negligible additional costs, indicating that adaptive partitioning incurs little overhead.

\begin{table}[t]
\centering
\caption{Average runtime per sample (s) on LongVideoBench at a budget of 32 frames, decomposed by processing stage.}
\label{tab:efficiency}
\setlength{\tabcolsep}{3pt}
\renewcommand{\arraystretch}{1.12}
\small
\begin{tabular}{lccccc}
\toprule
Method & Feature & Partition & Allocation & Selection & Total \\
\midrule
AKS & 13.743 & $<0.001$ & -- & $<0.001$ & 13.744 \\
QCA & 14.280 & $<0.001$ & 0.002 & 0.009 & 14.291 \\
\rowcolor{orange!25}
\textbf{APS} & 13.743 & 0.045 & $<0.001$ & $<0.001$ & 13.789 \\
\bottomrule
\end{tabular}
\end{table}

\subsection{Generalization to Advanced LVLMs}
\label{sec:generalization}

To evaluate the transferability of APS to advanced commercial LVLMs, we evaluate APS with GLM-5.3-Flash~\cite{zeng2026glm} and DeepSeek-V4.1-Flash~\cite{xu2026deepseek} on LongVideoBench under a fixed 16-frame budget. As shown in Table~\ref{tab:advanced_lvlms}, APS consistently improves performance over uniform sampling on both models, with larger gains on longer videos (10--60 min), where preserving relevant visual evidence is more challenging under limited frame budgets. These results demonstrate that APS remains effective with advanced commercial LVLMs beyond the backbones used in the main experiments.

\begin{table}[t]
\centering
\small
\caption{Generalization to advanced LVLMs on LongVideoBench with a 16-frame budget. $\Delta$ denotes the absolute accuracy improvement over uniform sampling in percentage points.}
\label{tab:advanced_lvlms}
\setlength{\tabcolsep}{4pt}
\begin{tabular}{llccc}
\toprule
\textbf{Model} & \textbf{Duration} & \textbf{Uniform} & \textbf{APS} & $\boldsymbol{\Delta}$ \\
\midrule
\multirow{3}{*}{GLM-5.3-Flash}
& 0--10 min  & 68.35 & \textbf{73.28} & +4.93 \\
& 10--60 min & 53.53 & \textbf{61.66} & +8.13 \\
& Overall    & 62.08 & \textbf{68.36} & \textbf{+6.28} \\
\midrule
\multirow{3}{*}{DeepSeek-V4.1-Flash}
& 0--10 min  & 67.83 & \textbf{71.73} & +3.90 \\
& 10--60 min & 54.42 & \textbf{60.42} & +6.00 \\
& Overall    & 62.15 & \textbf{66.94} & \textbf{+4.79} \\
\bottomrule
\end{tabular}
\end{table}

%% file: sec/5_conclusion.tex
\section{Conclusion}
\label{sec:conclusion}

In this work, we presented APS, a training-free framework for long-video frame selection. APS formulates temporal partitioning as a complexity-regularized fitting problem, enabling adaptive boundary determination. Experiments on three benchmarks demonstrate consistent gains over uniform sampling and competitive performance across diverse LVLMs. Further analyses validate its key components, robustness across diverse evaluation settings, and transferability to advanced LVLMs.